\documentclass[letterpaper, 10 pt, conference]{ieeeconf}  

\IEEEoverridecommandlockouts                              

\usepackage{amsmath,amssymb,amsfonts}
\usepackage{algpseudocode,algorithm}
\usepackage{subcaption}
\usepackage{graphicx}
\usepackage{cite}

\title{\LARGE \bf
L2C2: Locally Lipschitz Continuous Constraint \\towards Stable and Smooth Reinforcement Learning
}

\author{Taisuke Kobayashi$^{1}$
\thanks{$^{1}$Taisuke Kobayashi is with the Division of Information Science, Nara Institute of Science and Technology, 8916-5 Takayama-cho, Ikoma, Nara 630-0192, Japan
        {\tt\small kobayashi@is.naist.jp}}%
}

\begin{document}

\maketitle
\thispagestyle{empty}
\pagestyle{empty}

\begin{abstract}

This paper proposes a new regularization technique for reinforcement learning (RL) towards making policy and value functions smooth and stable.
RL is known for the instability of the learning process and the sensitivity of the acquired policy to noise.
Several methods have been proposed to resolve these problems, and in summary, the smoothness of policy and value functions learned mainly in RL contributes to these problems.
However, if these functions are extremely smooth, their expressiveness would be lost, resulting in not obtaining the global optimal solution.
This paper therefore considers RL under local Lipschitz continuity constraint, so-called L2C2.
By designing the spatio-temporal locally compact space for L2C2 from the state transition at each time step, the moderate smoothness can be achieved without loss of expressiveness.
Numerical noisy simulations verified that the proposed L2C2 outperforms the task performance while smoothing out the robot action generated from the learned policy.

\end{abstract}


\section{Introduction}

As the workforce shrinks and robotic technology develops, the tasks required of robots becomes more complex: e.g.
physical human-robot interaction~\cite{modares2015optimized,kobayashi2021whole};
work on disaster sites~\cite{kobayashi2015selection,delmerico2019current};
and manipulation of various objects~\cite{tsurumine2019deep,kroemer2021review}.
Many of these complex tasks are difficult to design in advance, and reinforcement learning (RL)~\cite{sutton2018reinforcement} has received a lot of attention as an alternative to classic model-based control.
In particular, deep RL, which approximates policy and value functions learned inside RL by deep neural networks (DNNs)~\cite{lecun2015deep}, has high expressiveness and potential to be applied to various problems.

In the last decade, there has been a great leap forward in the development of RL theory and verified in numerical simulations.
However, a large gap is still remained for the success of RL in the real world.
For example, RL requires a large number of samples, which cannot be efficiently collected for real-world problems.
Learning the policy robust to the simulation-real gap in massive parallel simulations has been promoted~\cite{rudin2021learning}, but it is not possible to resolve the problems with no simulators.
In the case of learning only with samples from real robots, it is prone to overfit because the number of data is small compared to the number of DNNs parameters~\cite{cobbe2019quantifying}, causing unstable learning.
The unpredictable behaviors by the learned policy would also be a critical issue, namely, the lacks of safety~\cite{achiam2017constrained,thananjeyan2021recovery} and smoothness~\cite{mysore2021train} must be resolved as auxiliary objectives in real robotic control applications.

Among the above many issues, this paper focuses on the smoothness of the policy and value functions in RL.
Theoretically, both functions should be arbitrary in shape, i.e. the discrete change should be representable.
In practical use, however, it is rare to jump them discretely with a slight change in state.
We can expect that both the policy and value functions are smooth.
In addition, their smoothness would prevent overfitting and stabilize the learning process.
However, function approximation by DNNs does not guarantee such smoothness, especially with small datasets.

Two related papers to achieve the smoothness are briefly introduced here.
For the value function, its temporal smoothness is easily expected, and Thodoroff et al.~\cite{thodoroff2018temporal} proposed a regularization method that treats this smoothness explicitly by adding a moving average of past value functions to the target value.
For the policy function, conditioning for action policy smoothness (CAPS)~\cite{mysore2021regularizing} directly regularizes the policy to chain the ones at adjacent time steps and to constrain the one over the properly random state (see later).

The common issue to these previous studies is the performance degradation caused by excessive regularization.
In short, over-smoothing the policy and value functions impairs the expressiveness that is originally required for accomplishing tasks.
In CAPS paper, the relation to Lipschitz continuity has been pointed out, but if it is enough constrained, the policy will lose its state-dependent adaptive behaviors.
With tough hyperparameters tuning, this problem can be mitigated in an ad-hoc manner, but a new design that does not allow for excessive regularization is desired.

\begin{figure}[tb]
    \centering
    \includegraphics[keepaspectratio=true,width=0.96\linewidth]{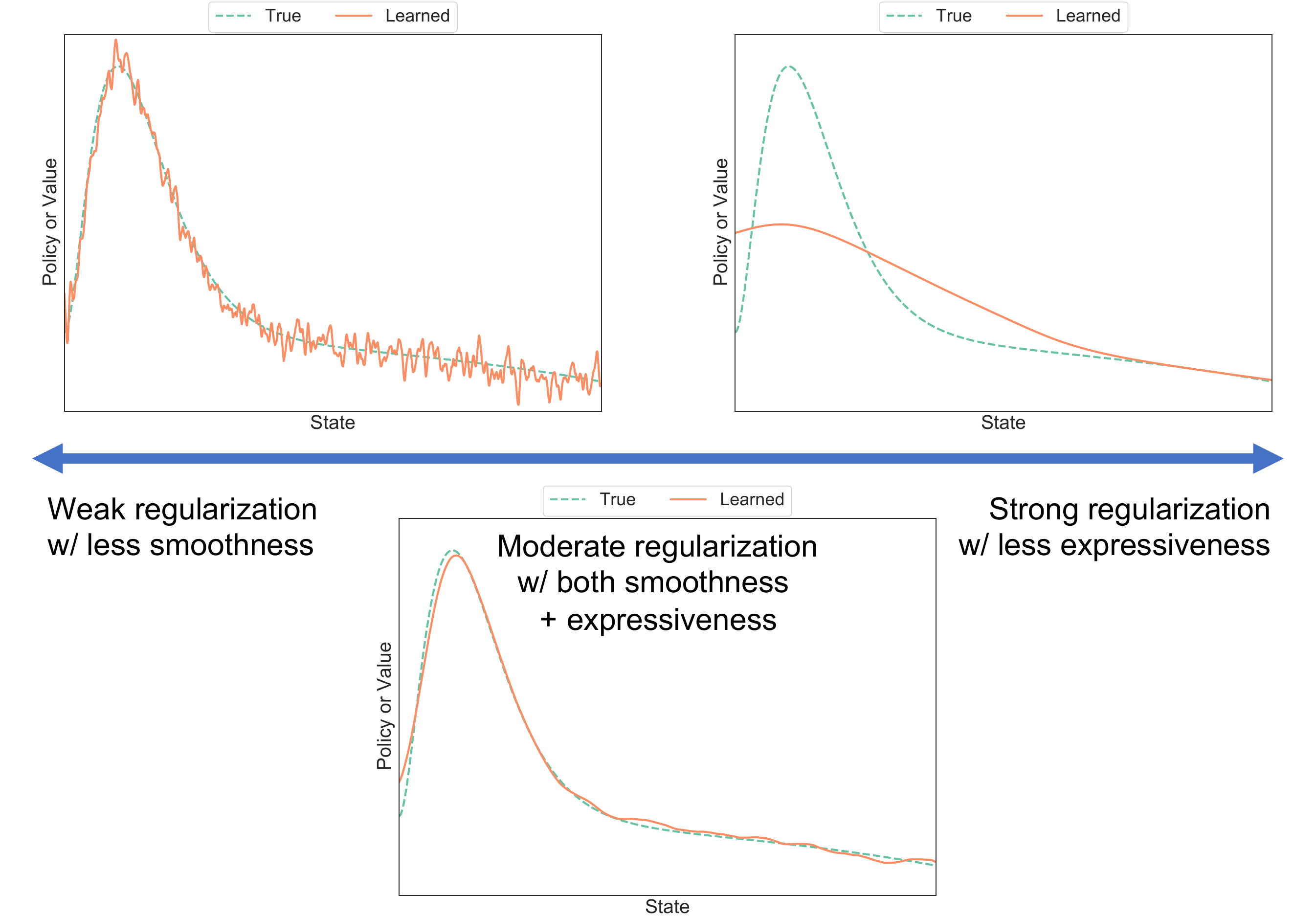}
    \caption{Conceptual sketches of the proposed method:
    function approximation by DNNs does not guarantee the smoothness of the learned function, inducing the behavior fluctuation with slight state changes;
    when regularization is applied to the entire state space excessively, the learned function would fail to capture the true function;
    by extracting local state spaces for the regularization, the smoothness can be obtained without losing the expressiveness of the function.
    }
    \label{fig:concept}
\end{figure}

Hence, this paper designs a new regularization method for making the policy and value functions moderate smooth (see Fig.~\ref{fig:concept}).
To this end, instead of \textit{global} Lipschitz continuity, the proposed method aims to satisfy \textit{local} Lipschitz continuous constraint, so-called L2C2.
The locally compact space for L2C2 can be determined spatio-temporally according to state transition at each time step.
Specifically, the locally compact space is defined by a box with the state at the pre-transition as the center and the state at post-transition as one of the vertices.
In such locally compact space, the policy and value functions are regularized to satisfy L2C2.
This design explicitly limits the range of smoothing and maintains sufficient expressiveness since adjacent states follow different L2C2.

The proposed L2C2 is verified with four simulation tasks in Pybullet with white noise.
The results show that L2C2 achieves the equivalent or better performance compared to the conventional methods without losing the expressiveness of the policy.
In addition, L2C2 succeeds in smoothing the policy better than the case without regularization.

\section{Preliminaries}

\begin{figure*}[tb]
    \begin{subfigure}[b]{0.32\linewidth}
        \centering
        \includegraphics[keepaspectratio=true,width=\linewidth]{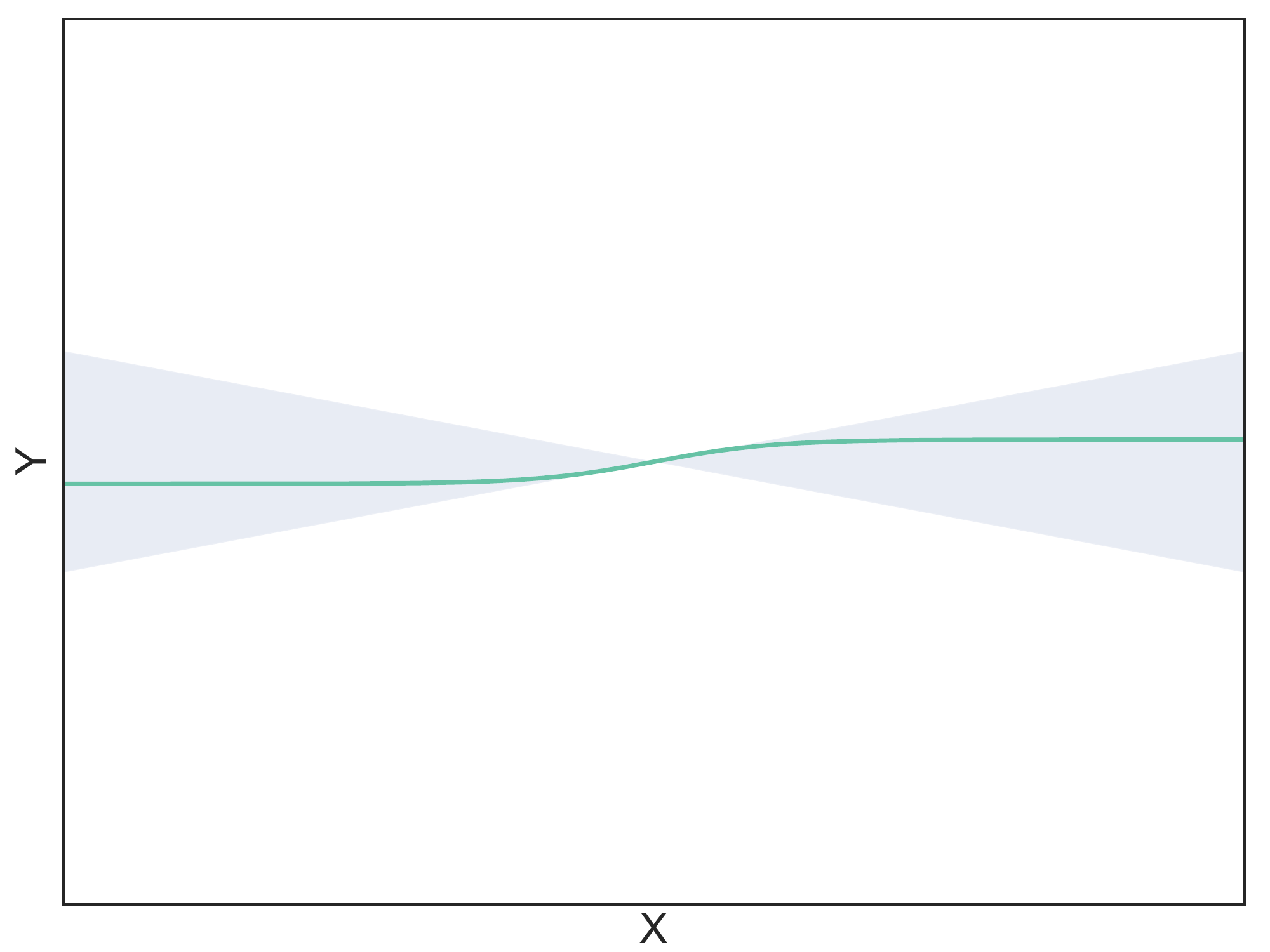}
        \subcaption{With small $K=1$}
        \label{fig:lipschitz_small}
    \end{subfigure}
    \begin{subfigure}[b]{0.32\linewidth}
        \centering
        \includegraphics[keepaspectratio=true,width=\linewidth]{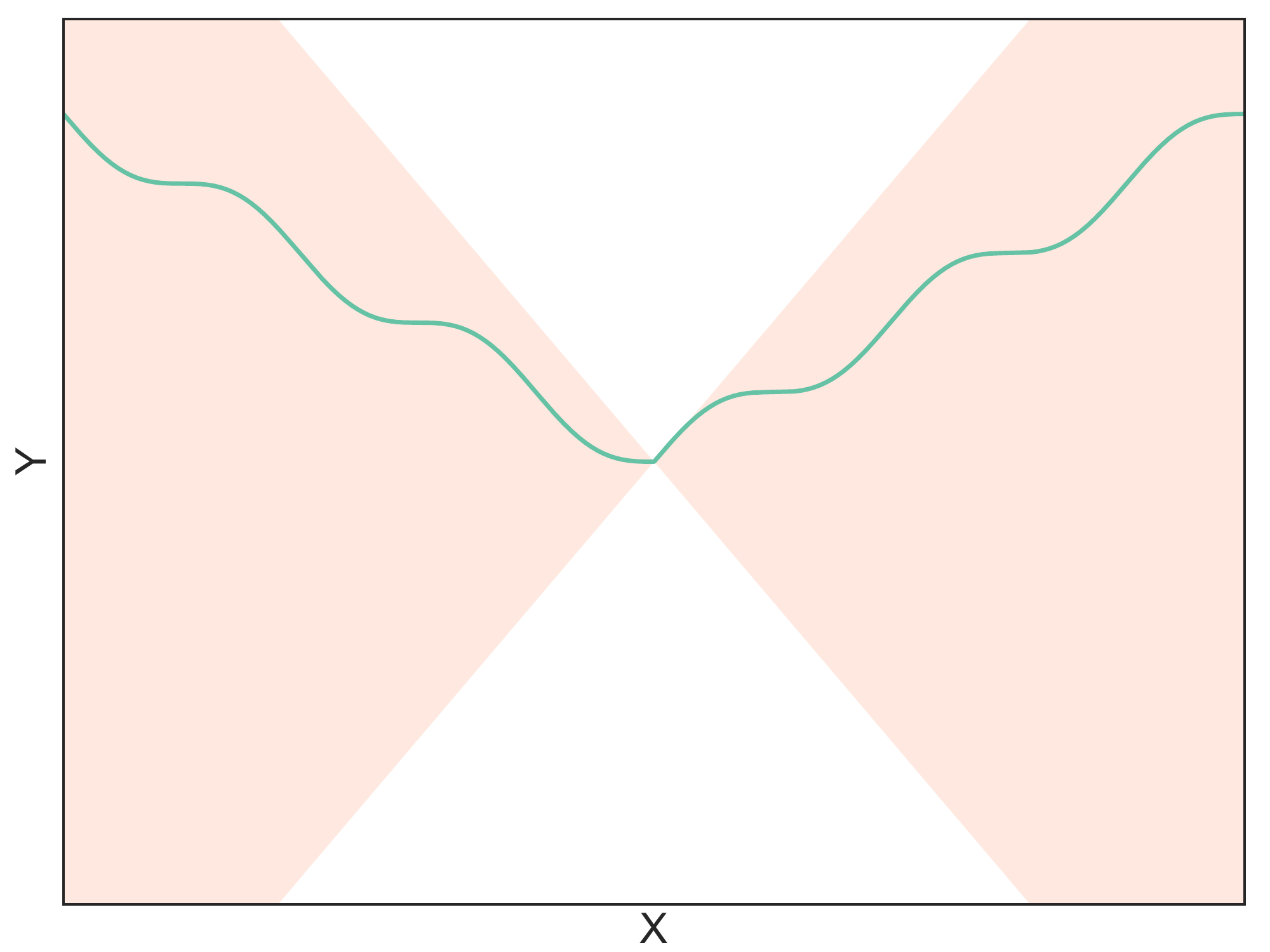}
        \subcaption{With large $K=2\pi$}
        \label{fig:lipschitz_large}
    \end{subfigure}
    \begin{subfigure}[b]{0.32\linewidth}
        \centering
        \includegraphics[keepaspectratio=true,width=\linewidth]{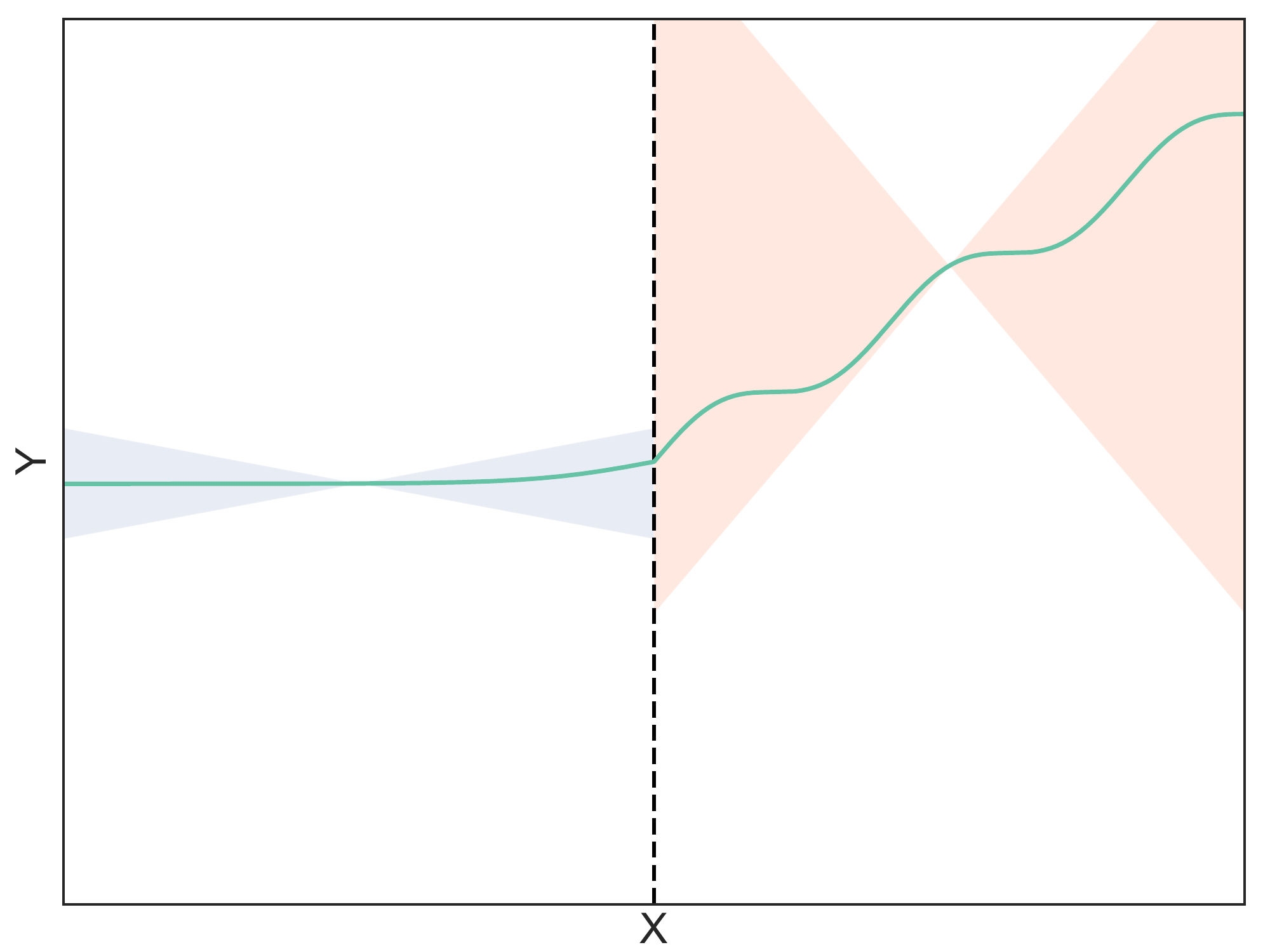}
        \subcaption{With small $K_1=1$ and large $K_2=2\pi$}
        \label{fig:lipschitz_local}
    \end{subfigure}
    \caption{Examples of Lipschitz continuous functions:
    global Lipschitz continuity in (a) and (b), large $K$ yields high expressiveness, while small $K$ makes the function smooth;
    local Lipschitz continuity like (c) allows each region to have a different $K$, so that expressiveness and smoothness can be selectively given to each part of the function.
    }
    \label{fig:lipschitz}
\end{figure*}

\subsection{Reinforcement learning}

The baseline RL algorithm employed in this paper is briefly introduced.
In robot control, actions often take continuous values, hence an actor-critic algorithm, which can handle continuous action space, is natural choice~\cite{sutton2018reinforcement}.
Note that the proposed method can be applied to other algorithms, such as Q-learning suitable for discrete action space.

In the RL framework, an agent interacts with an unknown environment under Markov decision process (MDP).
Specifically, the environment provides the current state $s$ from its initial randomness, $p_0(s)$, or its state transition probability, $p_e(s^\prime \mid s, a)$ and $s \gets s^\prime$.
Here, $s^\prime$ denotes the next state, which is stochastically determined according to $s$ and the agent's action $a$.
Therefore, since the agent can act on the environment's state transition through $a$, the goal is to find the desired state by acting appropriately.
To this end, $a$ is sampled from a state-dependent trainable policy function, $\pi(a \mid s; \theta_{\pi})$, with its parameters set $\theta_{\pi}$ (e.g. weights and biases of DNNs).
The outcome of the state transition is communicated to the agent as reward, $r = r(s, a, s^\prime)$.
RL maximizes the sum of $r$ over the future (so-called return), $R = \sum_{k=0}^\infty \gamma^k r_k$, with $\gamma \in [0, 1)$ discount factor and $k$ step ahead.
$R$ cannot be gained at the current time step, hence, its expected value is often inferred as a trainable (state) value function, $V(s; \theta_{V}) \simeq \mathbb{E}[R \mid s]$, with its parameters set $\theta_{V}$.
In these problem setting, RL optimizes the policy $\pi$ to maximize the correct value $V$ for arbitrary states.

To learn $V$ at first, a temporal difference (TD) error method is widely used.
It utilizes the fact that $V$ can be approximately regarded as recurrence relation, and minimizes the error with respect to the target signal $y$ as follows:
\begin{align}
    \mathcal{L}_{V}(\theta_{V}) &= \cfrac{1}{2} (y - V(s; \theta_{V}))^2
    \label{eq:loss_value} \\
    y &= r + \gamma V(s^\prime; \bar{\theta}_{V})
\end{align}
where a target network with the parameters set, $\bar{\theta}_{V}$, which is slowly updated towards $\theta_{V}$ (see the literature~\cite{kobayashi2021t}), is often introduced to stabilize the update of DNNs.
By minimizing $\mathcal{L}_{V}$, $\theta_{V}$ can be optimized to correctly infer the value over $s$.

To learn $\pi$, a policy-gradient method can be applied.
In brief, the goodness of the action taken can be estimated from the value changed by it, and a negative log-likelihood weighted by such goodness can be utilized for the optimization of $\theta_{\pi}$ as follows:
\begin{align}
    \mathcal{L}_{\pi}(\theta_{\pi}) &= - (y - V(s; \theta_{V})) \ln \pi(a \mid s; \theta_{\pi})
    \label{eq:loss_policy}
\end{align}
By minimizing $\mathcal{L}_{\pi}$, $\theta_{\pi}$ can be optimized to reach the state with higher value.
As a remark, the above updates can be done with a tuple $(s, a, s^\prime, r)$ (and sometimes, likelihood of a sampler alternative to $\pi$ using importance sampling technique), hence experience replay like~\cite{schaul2015prioritized}, which stores the tuple and replays them internally, is often utilized to improve sample efficiency.

It has been pointed out that $\pi$ and $V$ above can be interpreted as a generator and a discriminator in a generative adversarial network (GAN)~\cite{pfau2016connecting}.
GAN is also widely known for learning instability, but it has been reported that making the discriminator Lipschitz continuous can improve its stability~\cite{miyato2018spectral}.
This fact is also the basis for considering the Lipschitz continuity of the value function in this paper.

\subsection{Conditioning for action policy smoothness: CAPS}

As a baseline of the proposed method, CAPS~\cite{mysore2021regularizing} is briefly introduced.
Note that its description is modified from original according to the proposed method in this paper.

CAPS has been proposed for making the policy $\pi$ smooth by spatial and temporal regularizations.
Specifically, the following is minimized in addition to eqs.~\eqref{eq:loss_value} and~\eqref{eq:loss_policy}.
\begin{align}
    \mathcal{L}_{\mathrm{CAPS}}(\theta_{\pi}) &=
    \lambda_T d_\mathrm{CAPS}(\pi(\cdot \mid s; \theta_{\pi}), \pi(\cdot \mid s^\prime; \theta_{\pi}))
    \nonumber \\
    &+ \lambda_S d_\mathrm{CAPS}(\pi(\cdot \mid s; \theta_{\pi}), \pi(\cdot \mid \tilde{s}; \theta_{\pi}))
    \label{eq:loss_caps} \\
    d_\mathrm{CAPS}(\pi_1, \pi_2) &= \cfrac{1}{2} \| \mu_1 - \mu_2 \|_2^2
    , \
    \tilde{s} \sim \mathcal{N}(s, \sigma)
    \nonumber
\end{align}
where $\lambda_{T,S}$ are gains of regularizations, $\mu$ denotes the mean of $\pi$, and $\sigma$ decides the randomness of $\tilde{s}$, which is sampled from normal distribution.

The first term suppresses the temporal variation of the policy, and the second term ideally leads to local smoothness of the policy.
In practice, however, $\tilde{s}$ is in the entire real space due to normal distribution, and it will give global smoothness if $\sigma$ is not set appropriately.
In addition, since the suppression of temporal variation is chained to the entire trajectory, it can be regarded as restricting $\pi$ to the vicinity of a certain stationary policy.
Anyway, CAPS is expected to improve the smoothness, but excessive regularization may lead to failure in learning the optimal policy.

\subsection{Lipschitz continuity}

This paper focuses on the (global) Lipschitz continuity as a notion of function smoothness.
Its definition is given as below.
Suppose two metric spaces, $(X, d_{X})$ and $(Y, d_{Y})$, with the sets $X$ and $Y$ and their distance metrics $d_{X}$ and $d_{Y}$, respectively.
With them, a function $f: X \to Y$ is called (global) Lipschitz continuous if the following inequality is satisfied for all $x_1, x_2 \in X$ .
\begin{align}
    d_{Y}(f(x_1), f(x_2)) \leq K d_{X}(x_1, x_2)
    \label{eq:def_glc}
\end{align}
where $K \geq 0$ is so-called Lipschitz constant.
Intuitively, $K$ is consistent with the smoothness of $f$, i.e. $f$ is smoother with smaller $K$, as can be seen in Figs.~\ref{fig:lipschitz_small} and~\ref{fig:lipschitz_large}.

In DNNs, the global Lipschitz continuity is always satisfied if the activation function is chosen to be continuously differentiable.
$K$ of DNNs is however unclear, although its estimation method has been proposed~\cite{scaman2018lipschitz}.
Since DNNs would be unstable if $K$ is too large, its regularization method has also been proposed~\cite{gouk2021regularisation}.
It should be noted that this regularization can constrain the global $K$ explicitly, but it is difficult to limit it locally as this paper aims to do.

A relaxation of the conditions from eq.~\eqref{eq:def_glc} is the local Lipschitz continuity.
This is the case where the Lipschitz continuity is satisfied in a neighborhood $U_i \subset X$ of $x_i \in X$, and can be defined as follows:
\begin{align}
    d_{Y}(f(x_1), f(x_2)) \leq K_i d_{U_i}(x_1, x_2)
    \label{eq:def_llc}
\end{align}
where $K_i \geq 0$ denotes Lipschitz constant for $i$-subset, $d_{U_i}$ denotes the distance metric for $U_i$, and $x_1, x_2 \in U_i$.
In this definition, therefore, different smoothness is allowed for each subset, as illustrated in Fig.~\ref{fig:lipschitz_local}.

\section{Proposal}

\subsection{Theoretical formulation of hard-constrained problem}

As mentioned above, CAPS as baseline considers the global Lipschitz continuous constraint (only for the policy), causing the loss of expressiveness.
In contrast, since the local Lipschitz continuity does not refer to smoothness outside the neighborhood, it avoids the loss of expressiveness caused by extreme smoothness constraint, and preserves adaptability to changes in state.
Furthermore, since the function is smooth in the neighborhood, the behavior would be robust to small changes in state due to noise.
Therefore, this paper considers the local Lipschitz continuous constraint (L2C2).

Specifically, the following hard-constrained minimization problem is considered.
\begin{align}
    \theta_{\pi}^*, \theta_{V}^* = \arg\min_{\theta_{\pi}, \theta_{V}}
    \mathcal{L}_{\pi}(\theta_{\pi})
    + \mathcal{L}_{V}(\theta_{V})
    \label{eq:loss_hard}
\end{align}
such that
\begin{align}
    \sup_{s_1, s_2 \in U_{s}} \cfrac{d_{\pi}(\pi(a \mid s_1; \theta_{\pi}), \pi(a \mid s_2; \theta_{\pi}))}{d_{U_{s}}(s_1, s_2)}
    &\leq K_{U_{s}, \pi}
    \label{eq:cns_policy} \\
    \sup_{s_1, s_2 \in U_{s}} \cfrac{d_{V}(V(s_1; \theta_{V}), V(s_2; \theta_{V}))}{d_{U_{s}}(s_1, s_2)}
    &\leq K_{U_{s}, V}
    \label{eq:cns_value}
\end{align}
where $U_{s}$ denotes the neighborhood of state $s$ and $d_{\pi, V, U_{s}}$ are respective distance metrics.
The design of them is discussed in the section~\ref{subsec:l2c2_impl}.

Theoretically, Lipschitz constants $K_{U_{s}, \pi}$ and $K_{U_{s}, V}$ should be given for all $s$ to achieve both expressiveness and smoothness.
In other words, $K_{U_{s}, \pi,V}$ should be small for subsets that require high smoothness, while ones should be large for subsets where small change in state would cause jump in the policy and/or value.
However, as a matter of course, it is intractable to tune all $K_{U_{s}, \pi,V}$ in advance for each task.
In addition, such a hard-constrained minimization problem is infeasible because computation of the actual Lipschitz constant of DNNs is well-known as NP-hard~\cite{scaman2018lipschitz}.

\subsection{Practical formulation with soft regularization}

To convert the hard-constrained minimization problem in eqs.~\eqref{eq:loss_hard}--\eqref{eq:cns_value} into the solvable problem, two relaxation techniques are utilized in practice.
At first, the upper bounds in the hard constraints cannot be given analytically.
Even if they are found numerically by a sampling-based search method, its computational cost is too expensive.
Hence, the following constraints, $C_{U_{s}, \pi}(\theta_{\pi})$ and $C_{U_{s}, V}(\theta_{V})$, are alternatively employed.
\begin{align}
    \eqref{eq:cns_policy} &\geq \mathbb{E}_{\tilde{s} \in U_{s}} \left [ \cfrac{d_{\pi}(\pi(a \mid s; \theta_{\pi}), \pi(a \mid \tilde{s}; \theta_{\pi}))}{d_{U_{s}}(s, \tilde{s})} \right ]
    =: C_{U_{s}, \pi}(\theta_{\pi})
    \label{eq:reg_policy} \\
    \eqref{eq:cns_value} &\geq \mathbb{E}_{\tilde{s} \in U_{s}} \left [ \cfrac{d_{V}(V(s; \theta_{V}), V(\tilde{s}; \theta_{V}))}{d_{U_{s}}(s, \tilde{s})} \right ]
    =: C_{U_{s}, V}(\theta_{V})
    \label{eq:reg_value}
\end{align}
where the expectation manipulation can be easily approximated by Monte Carlo method.
Note that the computational cost is reduced by limiting the state pairs from $(s_1, s_2)$ in eqs.~\eqref{eq:cns_policy} and~\eqref{eq:cns_value} to $(s, \tilde{s})$ since the policy and value over $s$ is already computed for the main optimization target.

Another relaxation is based on Lagrange multiplier.
As the recent machine learning technologies have done~\cite{schulman2017proximal,higgins2017beta}, the constraints in a constrained optimization problem can be converted into regularization terms by Lagrange multiplier (or linear weighted scalarization of multi-objective optimization problem).
As consequence, the proposed method with L2C2 finally solves the following minimization problem.
\begin{align}
    \theta_{\pi}^*, \theta_{V}^* &= \arg\min_{\theta_{\pi}, \theta_{V}}
    \mathcal{L}_{\pi}(\theta_{\pi})
    + \mathcal{L}_{V}(\theta_{V})
    \nonumber \\
    &+ \lambda_{\pi} C_{U_{s}, \pi}(\theta_{\pi})
    + \lambda_{V} C_{U_{s}, V}(\theta_{V})
    \label{eq:loss_l2c2}
\end{align}
where $\lambda_{\pi, V}$ denote gains for the respective regularizations.

\begin{figure}[tb]
    \centering
    \includegraphics[keepaspectratio=true,width=0.96\linewidth]{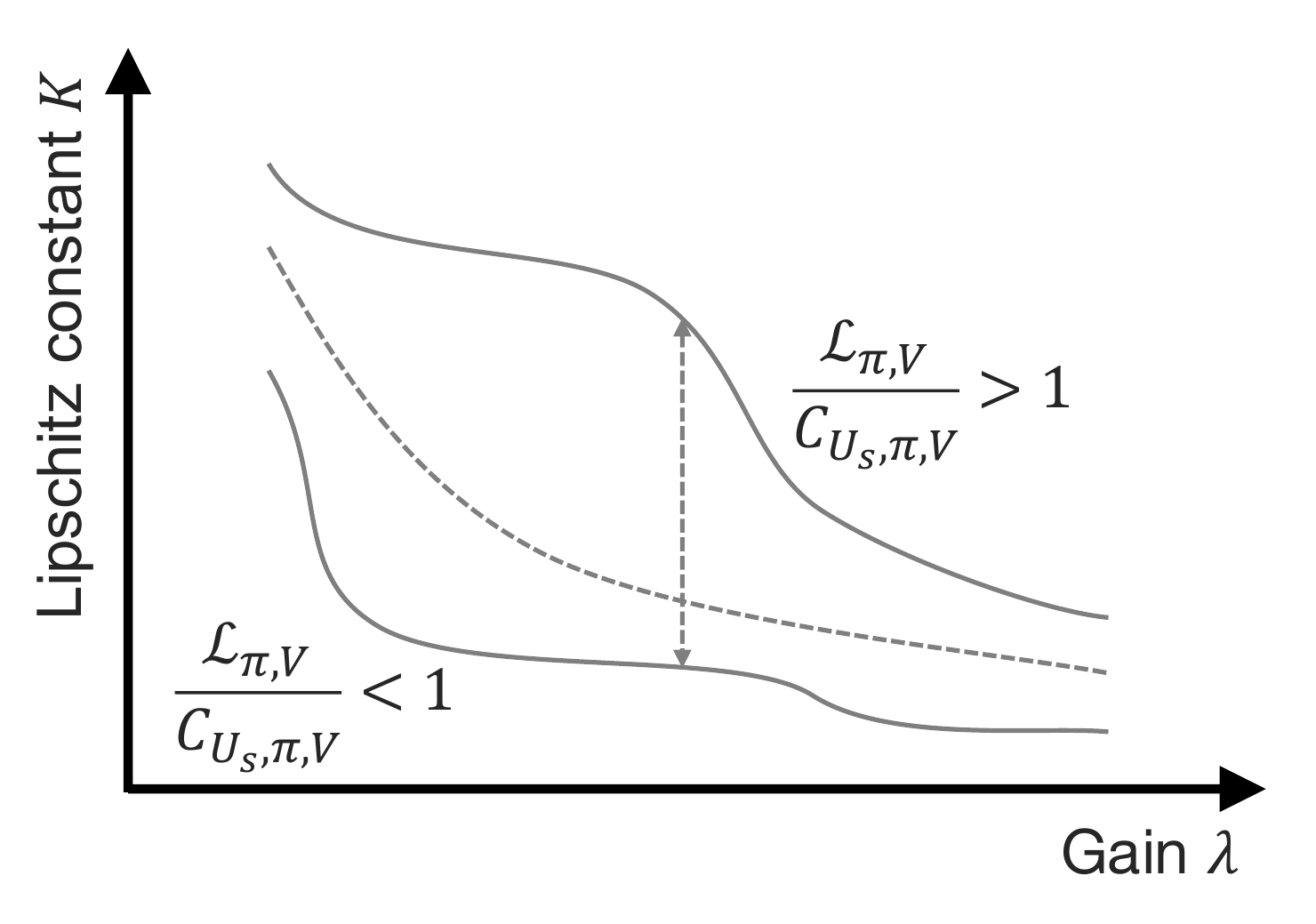}
    \caption{Image of implicitly adaptive constant $K_{U_{s}, \pi,V}$:
    even with the constant gain $\lambda_{\pi,V}$, the corresponding $K_{\pi,V}$ would be depending on the ratio of the main minimization target $\mathcal{L}_{\pi,V}$ and the regularization term $C_{U_{s}, \pi,V}$.
    }
    \label{fig:adaptiveness}
\end{figure}

As a remark, $\lambda_{\pi, V}$ are given to be constant for all the subsets, although the different Lipschitz constants $K_{U_{s}, \pi,V}$ are required in the hard-constrained minimization problem.
This is because, in the case of the regularization obtained through the above two relaxations, hard constraints are replaced by soft constraints.
Namely, when the original losses, $\mathcal{L}_{\pi}(\theta_{\pi})$ and $\mathcal{L}_{V}(\theta_{V})$, are small, there is room for satisfying the regularizations and thus the stronger constraints (i.e. the smaller $K_{U_{s}, \pi,V}$) are implicitly applied.
On the other hand, when $\mathcal{L}_{\pi}(\theta_{\pi})$ and $\mathcal{L}_{V}(\theta_{V})$ are large, which implies that large changes in the policy and value functions are required, the weaker constraints (i.e. the larger $K_{U_{s}, \pi,V}$) will naturally be expected.
As a result, $K_{U_{s}, \pi,V}$ for all the subsets can implicitly be adjusted depending on the main minimization targets, even with the constant $\lambda_{\pi, V}$ (see Fig.~\ref{fig:adaptiveness}).

\subsection{Implementation}
\label{subsec:l2c2_impl}

For the specific implementation, the abstracted factors in eq.~\eqref{eq:loss_l2c2}, i.e. the instance of $(U_{s}, d_{U_{s}})$, the way to sample $\tilde{s}$ from $U_{s}$, and the instance of $d_{\pi}$ and $d_{V}$, are designed.
It should be noted that the following design prioritizes ease of implementation and numerical stabilization, although there may be others that provide higher performance.

If the maximum distance to define the subset is given as a hyperparameter absolutely, it is hard to tune due to unknown range of possible states.
Therefore, the subset $U_{s}$ is temporally designed with the relative distance between $s$ and $s^\prime$ as follows:
\begin{align}
    U_{s}(s^\prime) = \left \{ \tilde{s} \mid \cfrac{d_{U_{s}}(s, \tilde{s})}{d_{U_{s}}(s, s^\prime)} \leq \sigma \right \}
    \label{eq:def_subset}
\end{align}
where $\sigma$ denotes the relative size of each subset and determines noise scale.
This design intuitively leads to the expectation that the policy and value functions during the state transition from $s$ to $s^\prime$ will change smoothly.

As for $d_{U_{s}}$, the dependency between the state dimensions, which is caused by like $L_2$ norm, would increase the cost of sampling.
The easiest way to avoid this problem, $L_{\infty}$ norm is natural choice.
In addition, $d_{U_{s}}$ plays an important role in the strength of the regularizations, as well as $U_{s}$, it is more stable to consider the relative (or normalized) distance based on the distance traveled $s \to s^\prime$.
From these requirements, the following distance is employed.
\begin{align}
    d_{U_{s}}(s, \tilde{s}; s^\prime) = \left \| \cfrac{\tilde{s} - s}{s^\prime - s} \right \|_{\infty} + \epsilon
    \label{eq:def_dis_state}
\end{align}
where $\epsilon > 0$ is added to avoid zero division, so strictly speaking, this design is regarded as an extended distance.

Thanks to the design of $d_{U_{s}}$ as $L_{\infty}$ norm, sampling of $\tilde{s}$ can be considered independently for each dimension.
In this paper, the proposed method samples $\tilde{s}$ from the simplest uniform distribution.
\begin{align}
    \tilde{s} &= s + (s^\prime - s) u
    \label{eq:def_sample} \\
    u &\sim \mathcal{U}(-\sigma - (\sigma - 1) \epsilon, \sigma + (\sigma - 1) \epsilon)
    \nonumber
\end{align}
where $\mathcal{U}(\underline{u}, \overline{u})$ denotes the uniform distribution of $(\underline{u}, \overline{u})$.

\begin{figure}[tb]
    \centering
    \includegraphics[keepaspectratio=true,width=0.96\linewidth]{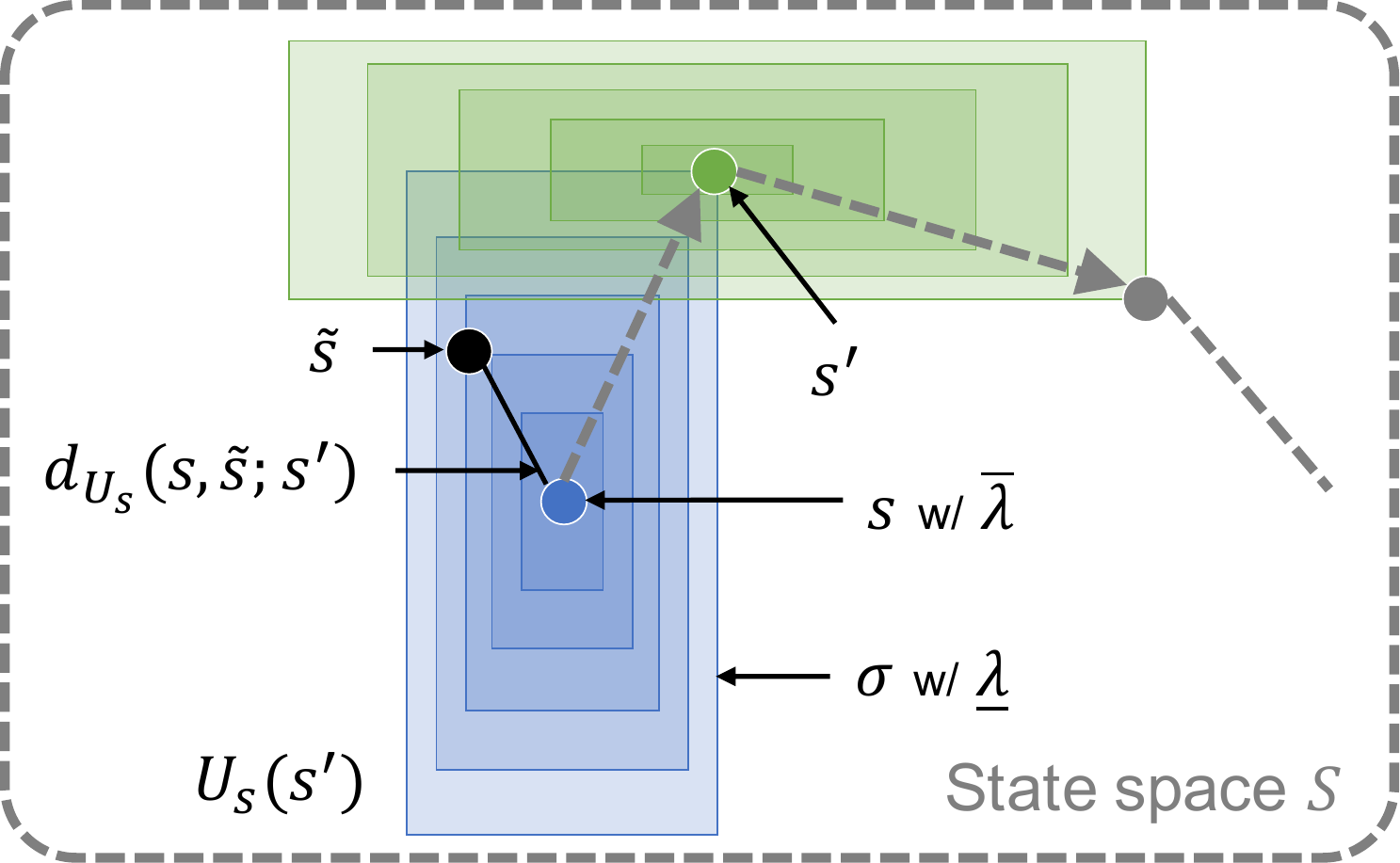}
    \caption{Geometry of the implementation:
    the neighborhood of $s$ is given by the state transition from $s \to s^\prime$ as $U_{s}(s^\prime)$ with its scale $\sigma$;
    the distance between $s$ and $\tilde{s} \sim U_{s}(s^\prime)$, $d_{U_{s}}(s, \tilde{s}; s^\prime)$, is measured as normalized $L_{\infty}$ norm;
    when $\tilde{s} \simeq s$, the adaptive gain $\tilde{\lambda}$ will be its upper bound $\overline{\lambda}$;
    when $\tilde{s}$ is close to the edge of the subset, $\tilde{\lambda}$ will be $\underline{\lambda}$.
    }
    \label{fig:implementation}
\end{figure}

Thanks to the existence of $\epsilon$, the range of $\tilde{\lambda}_{\pi, V}(s, \tilde{s}; s^\prime) = \lambda_{\pi, V} / d_{U_{s}}(s, \tilde{s}; s^\prime)$ (i.e. the distance-dependent gains for the respective regularizations) can be specified.
Specifically, given $\tilde{\lambda}_{\pi} \in (\underline{\lambda}, \overline{\lambda})$ with $\underline{\lambda} < \overline{\lambda}$ lower and upper bounds,
$\epsilon$ and $\lambda_{\pi}$ are determined as below.
\begin{align*}
    \epsilon = \cfrac{\sigma \underline{\lambda}}{\overline{\lambda} - \sigma \underline{\lambda}}
    , \
    \lambda_{\pi} = \overline{\lambda} \epsilon
\end{align*}
In addition, given $\lambda_{V} = \beta \lambda_{\pi}$ with $\beta > 0$, $\tilde{\lambda}_{V} \in (\beta \underline{\lambda}, \beta \overline{\lambda})$ can be specified.
The above implementation can be illustrated as shown in Fig.~\ref{fig:implementation}.

Finally, $d_{\pi}$ and $d_{V}$ should be designed in consideration of the fact that $\theta_{\pi}$ and $\theta_{V}$ are updated using their gradients.
To stabilize the updates, the squared error, which is with continuous gradient, would be effective.
Noting that $\pi$ is in the probability space, $d_{\pi}$ and $d_{V}$ are specified as follows:
\begin{align}
    &d_{\pi}(\pi(a \mid s; \theta_{\pi}), \pi(a \mid \tilde{s}; \theta_{\pi}))
    \nonumber \\
    &= \cfrac{1}{2} \int \left (\sqrt{\pi(a \mid s; \theta_{\pi})} - \sqrt{\pi(a \mid \tilde{s}; \theta_{\pi})} \right )^2 da
    \label{eq:def_dis_policy} \\
    &d_{V}(V(s; \theta_{V}), V(\tilde{s}; \theta_{V}))
    = \cfrac{1}{2} \| V(s; \theta_{V}) - V(\tilde{s}; \theta_{V}) \|^2_2
    \label{eq:def_dis_value}
\end{align}
$d_{\pi}$ is given as squared Hellinger distance, and its integral can be approximated by Monte Carlo method with importance sampling: i.e. $\int f(x) dx = \mathbb{E}_{p(x)}[f(x)/p(x)]$.
Note that, in the definition of $d_{V}$, $V$ is given as a vector assuming ensemble learning (e.g.~\cite{osband2016deep}).
If the action value function $Q(s, a)$ want to be trained, its distance can also be given as the squared error over $a$.

\subsection{Difference from CAPS}

While both the proposed method and CAPS have several points in common, such as the addition of regularization terms to increase the smoothness, several significant differences can also be confirmed.
In this section, the values of the proposed method is revealed by clarifying the important differences between the proposed method and CAPS.

The first difference is that CAPS regularizes the policy regardless of the distance from $s$, while the proposed method has the distance-dependent regularizations.
Therefore, for $\tilde{s} \simeq s$, where more smoothness is required, the proposed method can be sufficiently smooth without increasing the regularization gain, while CAPS requires a larger gain.
On the other hand, for distant states, the proposed method naturally weakens the regularizations without loss of expressiveness, while CAPS restricts the policy with equal strength, making tuning a well-balanced gain difficult.

Secondly, while CAPS gives spatial and temporal regularization independently, the proposed method combines spatio-temporal regularizations by designing neighborhoods (subsets) based on the state transitions.
In particular, the temporal regularization of CAPS may converge to a state-independent stationary policy in the extreme case.
In the proposed method, however, $s^\prime$ is near the boundary of the subset of $s$, so the distance-dependent regularizations described above would weaken the restriction and leave room for change.
In addition, when the regularizations for the subset of $s^\prime$ are performed, the regularization gain for $\tilde{s} \simeq s^\prime$ is stronger, preventing convergence to the stationary policy.

Another benefit from the spatio-temporal regularizations is that the relative regularization range eases tuning.
In CAPS, the noise scale to generate $\tilde{s}$ is specified absolutely, which inherently requires fine-tuning according to the faced problem.
In contrast, it in the proposed method is determined based on the distance traveled at each state transition, so that differences in the state space of each problem are not critical.

Finally, only in the proposed method, the regularization to the value function is added.
It is obvious that if the value fluctuates greatly with a slight change in state, the TD error will become noisy and the learning will become unstable.
In fact, the deep RL introduces the target networks to stabilize the TD error~\cite{kobayashi2021t}, suggesting that the smoothness of the value function is important.
The literature~\cite{fujimoto2018addressing} has also reported that the stability of learning can be improved by the smoothness of the value function, which shows the validity of the proposed method.

Although there are other minor differences, the characteristic differences above, summarized in Table~\ref{tab:diff}, allow the proposed method to obtain the policy and value functions that are sufficiently smooth locally without loss of expressiveness globally.
Such a moderate smoothness will provide stability in learning and robustness to noise, although its numerical verification should be shown later.

As a remark, state-adversarial MDP (SA-MDP)~\cite{zhang2020robust,zhang2021robust} has been studied with a similar but different purpose to the proposed method and CAPS.
In SA-MDP, a disturber intentionally adds adversarial noise in an ideal environment, where the true state is observable, to make the policy robust.
After training in the ideal environment, the learned policy is deployed to the corresponding real environment, the noise of which should be included in the adversarial noise.
In the proposed method and CAPS, the true state is unobservable, hence learning is based on state observations mixed with noise signals, which is a more realistic setting.

\begin{table}[tb]
    \caption{Characteristic differences}
    \label{tab:diff}
    \centering
    \begin{tabular}{ccc}
        \hline\hline
         & CAPS~\cite{mysore2021regularizing} & Proposed L2C2
        \\
        \hline
        Regularization gain & Constant & Distance-dependent
        \\
        Spatio-temporality & Independent & Unified
        \\
        Convergent policy & Stationary & State-dependent
        \\
        Noise scale & Absolute & Relative
        \\
        Value function & Not regularized & Regularized
        \\
        \hline\hline
    \end{tabular}
\end{table}

\section{Numerical verification}

\begin{table}[tb]
    \caption{Hyperparameters for the used RL algorithms}
    \label{tab:param}
    \centering
    \begin{tabular}{ccc}
        \hline\hline
        Symbol & Meaning & Value
        \\
        \hline
        $\gamma$ & Discount factor & $0.99$
        \\
        $(\alpha, \beta, \epsilon, \underline{\tilde{\nu}})$ & For AdaTerm~\cite{ilboudo2022adaterm} & $(10^{-3}, 0.9, 10^{-5}, 1.0)$
        \\
        $(\kappa, \beta, \lambda, \underline{\Delta})$ & For PPO-RPE~\cite{kobayashi2021proximal} & $(0.5, 0.5, 0.999, 0.1)$
        \\
        $(N_c, N_b, \alpha, \beta)$ & For PER~\cite{schaul2015prioritized} & $(10^4, 32, 1.0, 0.5)$
        \\
        $(\tau, \nu)$ & For t-soft update~\cite{kobayashi2021t} & $(0.1, 1.0)$
        \\
        \hline\hline
    \end{tabular}
\end{table}

\subsection{Setup}

For the statistical verification of the proposed method, the following simulations are conducted.
As simulation environments, Pybullet~\cite{coumans2016pybullet} with OpenAI Gym~\cite{brockman2016openai} is employed.
From it, \textit{InvertedPendulumSwingupBulletEnv-v0} (Swingup), \textit{ReacherBulletEnv-v0} (Reacher), \textit{HopperBulletEnv-v0} (Hopper), and \textit{AntBulletEnv-v0} (Ant) are chosen as tasks in this paper.
To make the tasks harder and more practical, the observations from them are with white noises, scale of which is $0.01$ for Swingup and Reacher and $0.001$ for Hopper and Ant (depending on their original difficulty).
With one random seed (18 seeds in total), each task is tried to be accomplished by each method.
The sum of rewards is computed for each episode as a score (larger is better).
$L_2$ norm of temporal actions (i.e. $\|\mu - \mu^\prime\|_2$) is also evaluated to show the policy smoothness (smaller is better).
After training, the learned policy is deployed to run 100 times for evaluating the final scores.

All the algorithms are implemented with PyTorch~\cite{paszke2017automatic} basically.
To approximate the policy and value functions, two-layered fully connected networks with 100 neurons and Layer Normalization~\cite{ba2016layer} for each are inserted as hidden layers, respectively.
As an activation function, $x \sigma_s(x)$ with $\sigma_s$ derivative of Squareplus~\cite{barron2021squareplus}, so-called Squish, is employed.
The stochastic policy function is modeled by diagonal normal distribution, which is the most popular model for continuous action space.
The mean of five outputs from the networks is regarded as the ensemble value function for stable learning.
Note that the networks up to the output layer are shared for the five outputs, as in the literature~\cite{osband2016deep}, which reduces memory costs.
The parameter sets of the above networks, $\theta_{\pi}$ and $\theta_{V}$, are updated by AdaTerm~\cite{ilboudo2022adaterm}, which is robust to wrong supervised signals that are likely to appear in RL.

The proposed L2C2 (and CAPS) only adds the regularization terms to the original RL, that is, the underlying RL algorithm can freely be designed.
In this paper, PPO-RPE~\cite{kobayashi2021proximal} is employed to stabilize learning, and integrated with prioritized experience replay (PER)~\cite{schaul2015prioritized} to accelerate learning.
The target networks for both the policy and value functions are constructed, and robustly updated by t-soft update~\cite{kobayashi2021t}.
Hyperparamters for these techniques are summarized in Table~\ref{tab:param}.

The following three methods are compared.
\begin{itemize}
    \item Vanilla without regularization
    \item CAPS with $(\sigma=0.2, \lambda_T=0.01, \lambda_S=0.05)$
    \item L2C2 with $(\sigma=1.0, \underline{\lambda}=0.01, \overline{\lambda}=1.0, \beta=0.1)$
\end{itemize}
Note that the hyperparameters for CAPS were tuned depending on the tasks in the original paper~\cite{mysore2021regularizing} but their recommended values could not be found.
Therefore, the values found in the official implementation were adopted.
$\underline{\lambda}$ of L2C2 is made to match the minimum gain of CAPS, but the distance-dependent regularization for the locally compact space ensures expressiveness even with the large $\overline{\lambda}$.
As discussed later, stabilization of the value function is basically achieved with the algorithms implemented as above.
For this reason, the regularization of the value function is set to be smaller than that of the policy (i.e. $\beta = 0.1 < 1$).

\subsection{Result}

\begin{figure*}[tb]
    \begin{subfigure}[b]{0.24\linewidth}
        \centering
        \includegraphics[keepaspectratio=true,width=\linewidth]{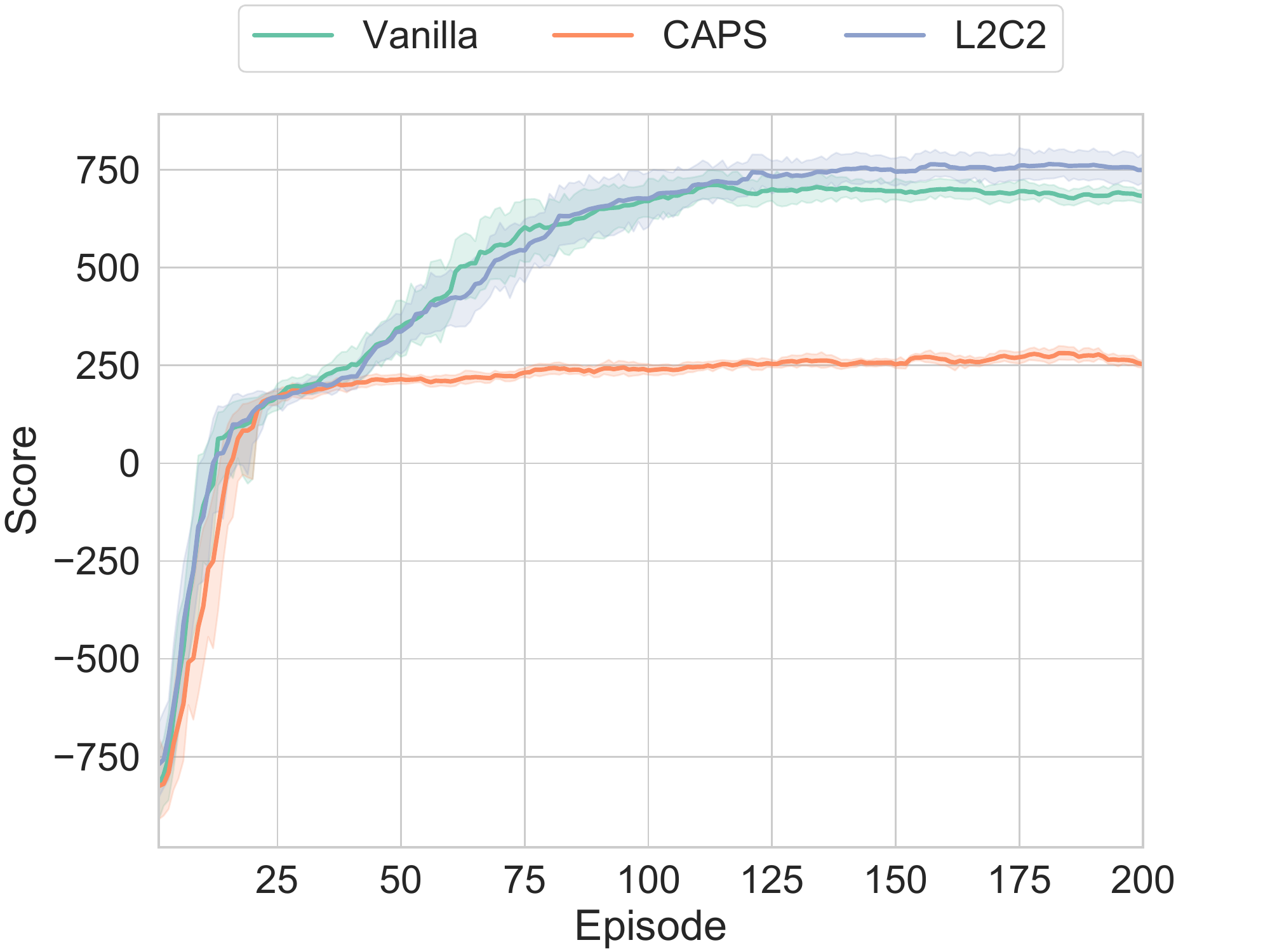}
        \subcaption{Swingup}
        \label{fig:score_Swingup}
    \end{subfigure}
    \begin{subfigure}[b]{0.24\linewidth}
        \centering
        \includegraphics[keepaspectratio=true,width=\linewidth]{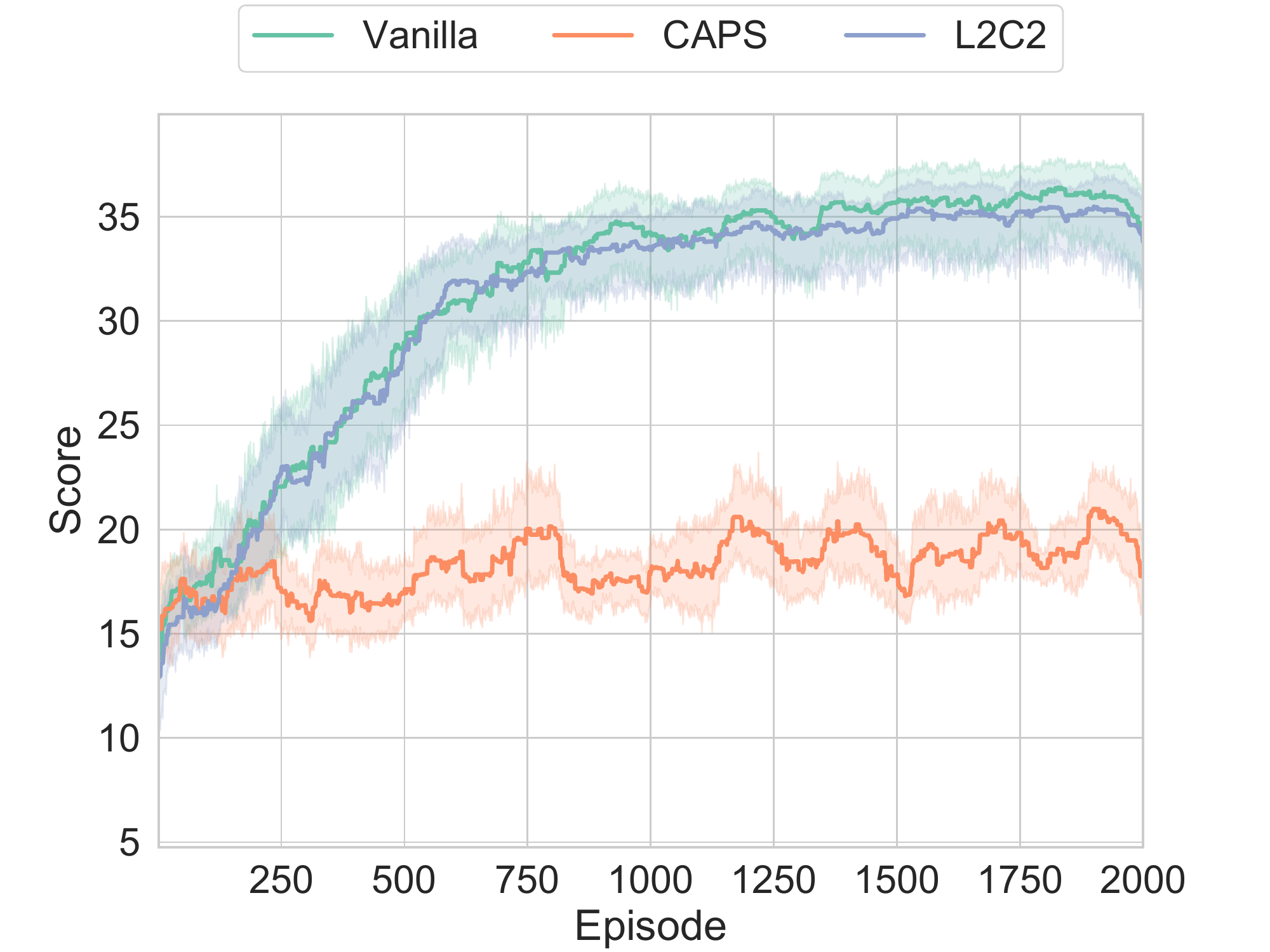}
        \subcaption{Reacher}
        \label{fig:score_Reacher}
    \end{subfigure}
    \begin{subfigure}[b]{0.24\linewidth}
        \centering
        \includegraphics[keepaspectratio=true,width=\linewidth]{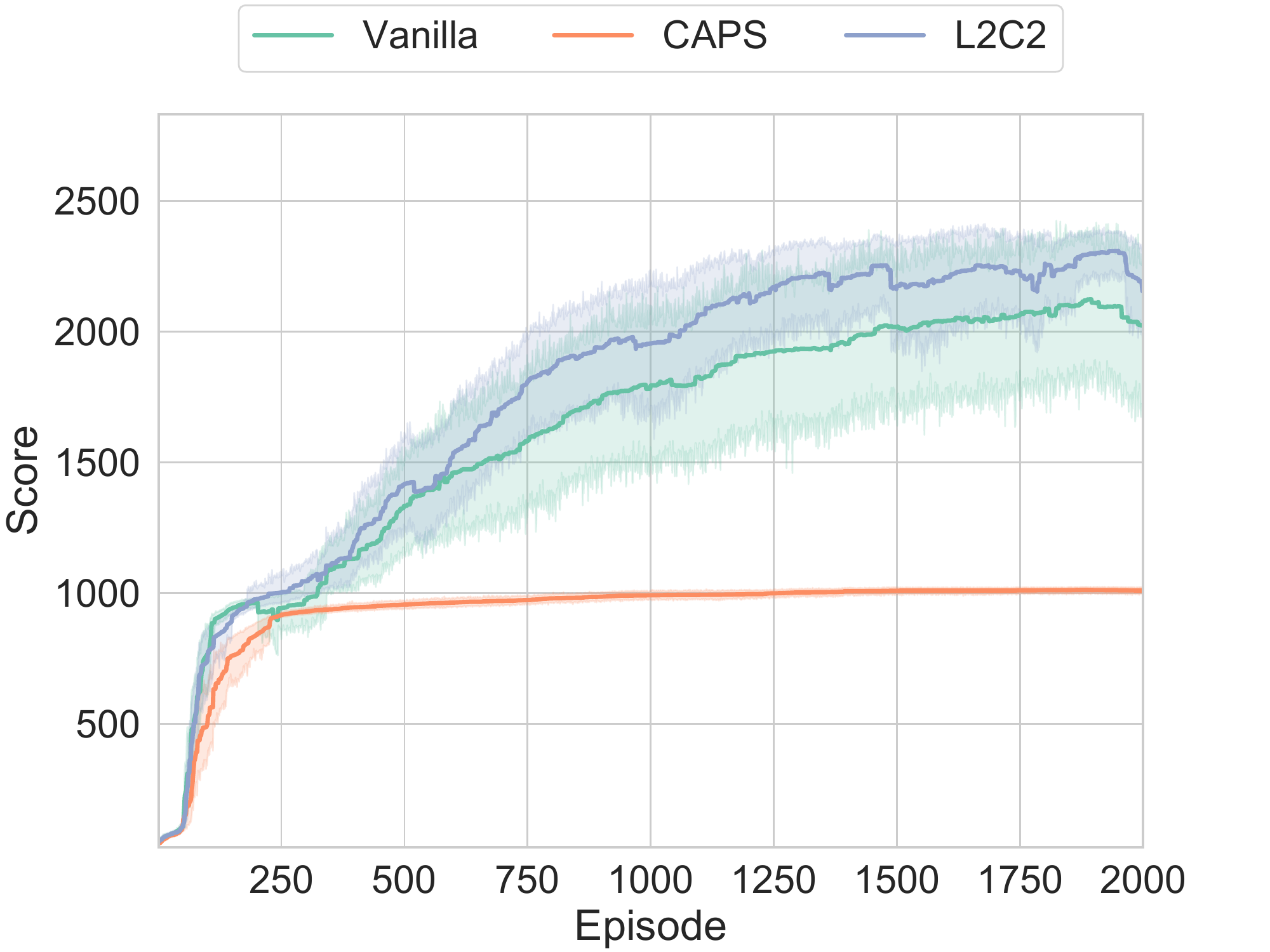}
        \subcaption{Hopper}
        \label{fig:score_Hopper}
    \end{subfigure}
    \begin{subfigure}[b]{0.24\linewidth}
        \centering
        \includegraphics[keepaspectratio=true,width=\linewidth]{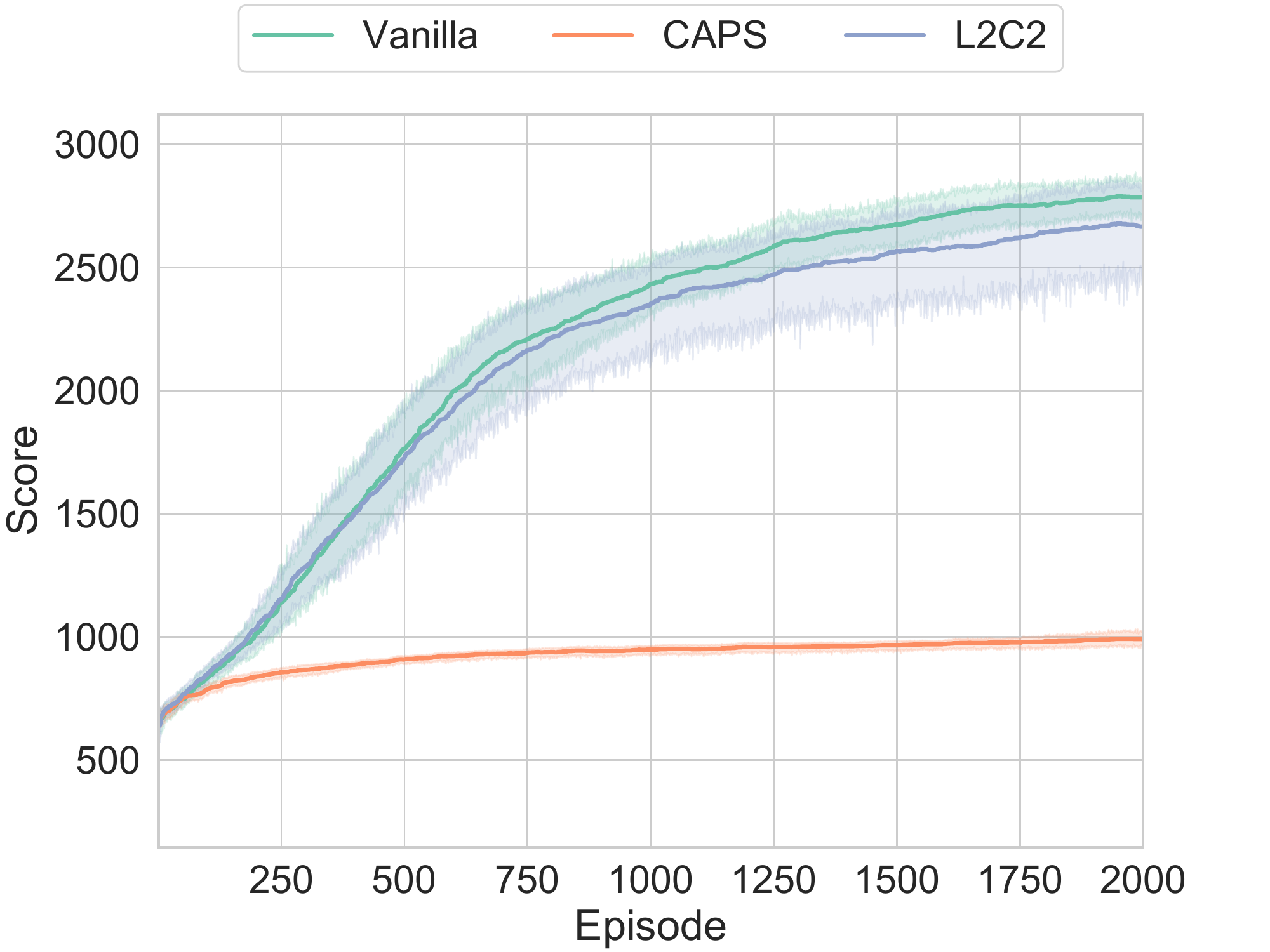}
        \subcaption{Ant}
        \label{fig:score_Ant}
    \end{subfigure}
    \caption{Learning curves for score (the sum of rewards):
    CAPS fell into local solutions in all the tasks;
    the proposed L2C2 achieved the equivalent or better learning performance compared to Vanilla;
    for (c) Hopper (the most unstable task in the benchmarks), L2C2 outperformed Vanilla by smoothing the policy;
    for (d) Ant with the highest-dimensional action space, sometimes, L2C2 would inhibited the exploration and delayed (or failed to) learning.
    }
    \label{fig:score}
\end{figure*}

\begin{figure*}[tb]
    \begin{subfigure}[b]{0.24\linewidth}
        \centering
        \includegraphics[keepaspectratio=true,width=\linewidth]{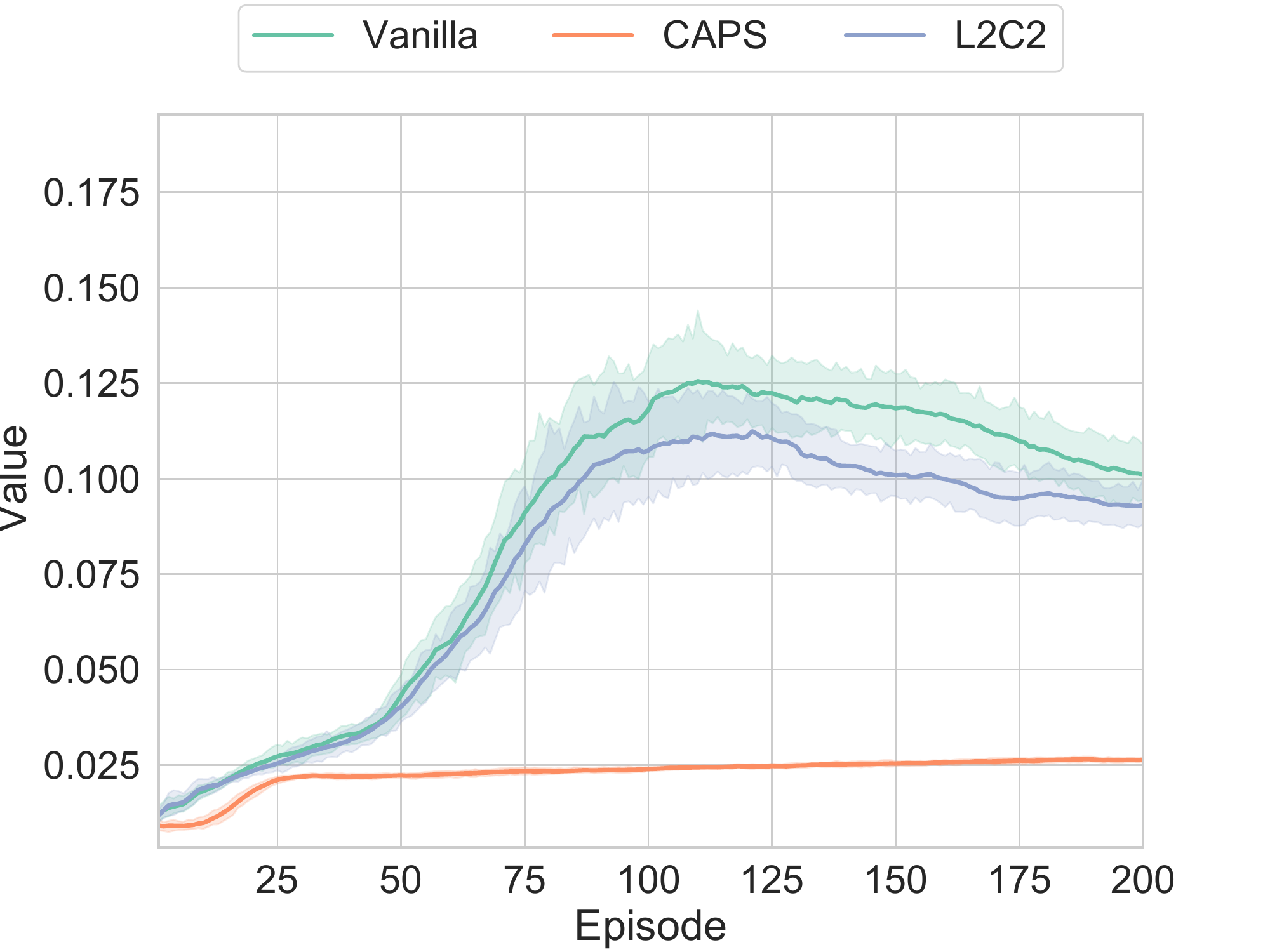}
        \subcaption{Swingup}
        \label{fig:smo_diff_action_Swingup}
    \end{subfigure}
    \begin{subfigure}[b]{0.24\linewidth}
        \centering
        \includegraphics[keepaspectratio=true,width=\linewidth]{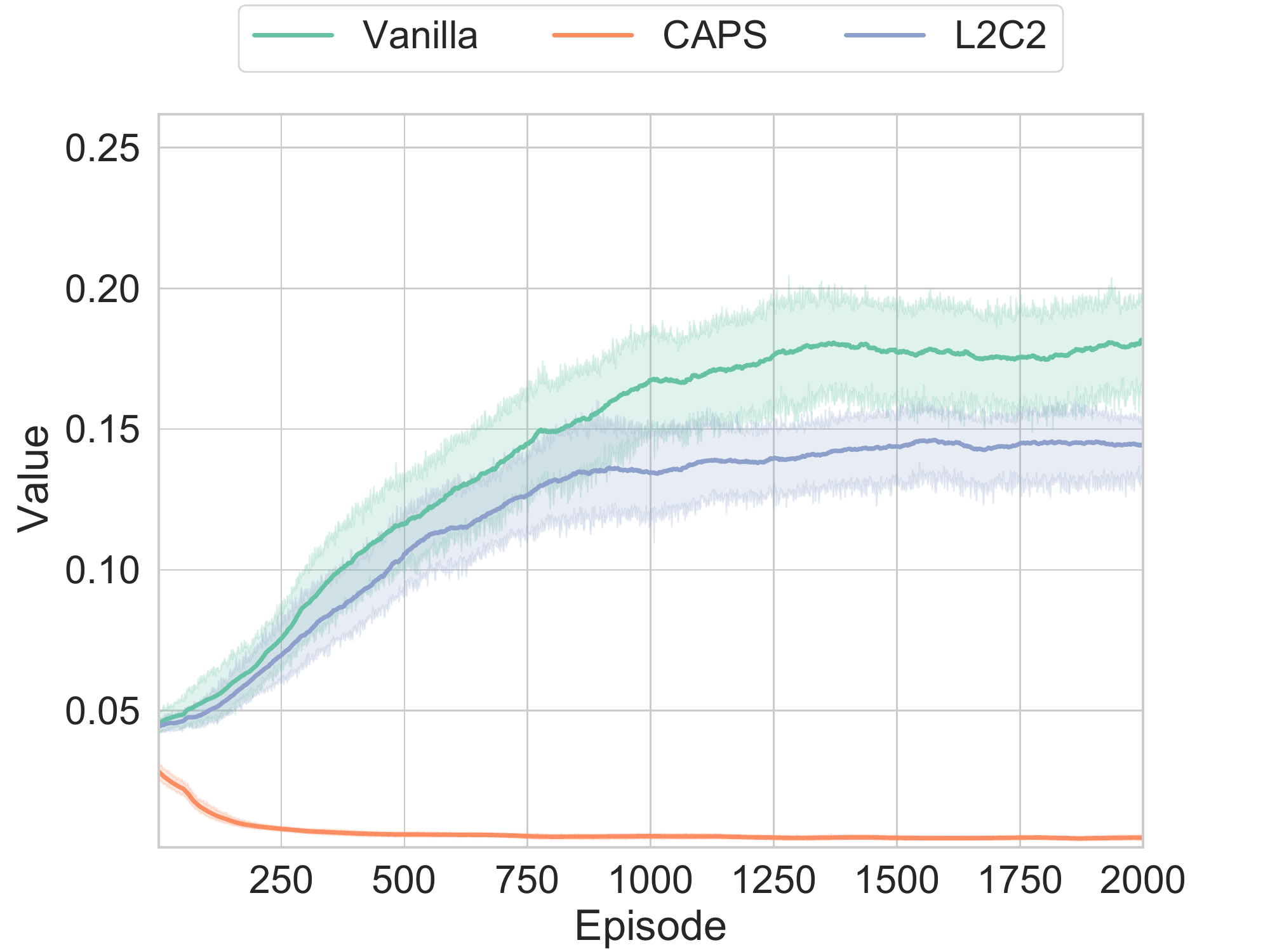}
        \subcaption{Reacher}
        \label{fig:smo_diff_action_Reacher}
    \end{subfigure}
    \begin{subfigure}[b]{0.24\linewidth}
        \centering
        \includegraphics[keepaspectratio=true,width=\linewidth]{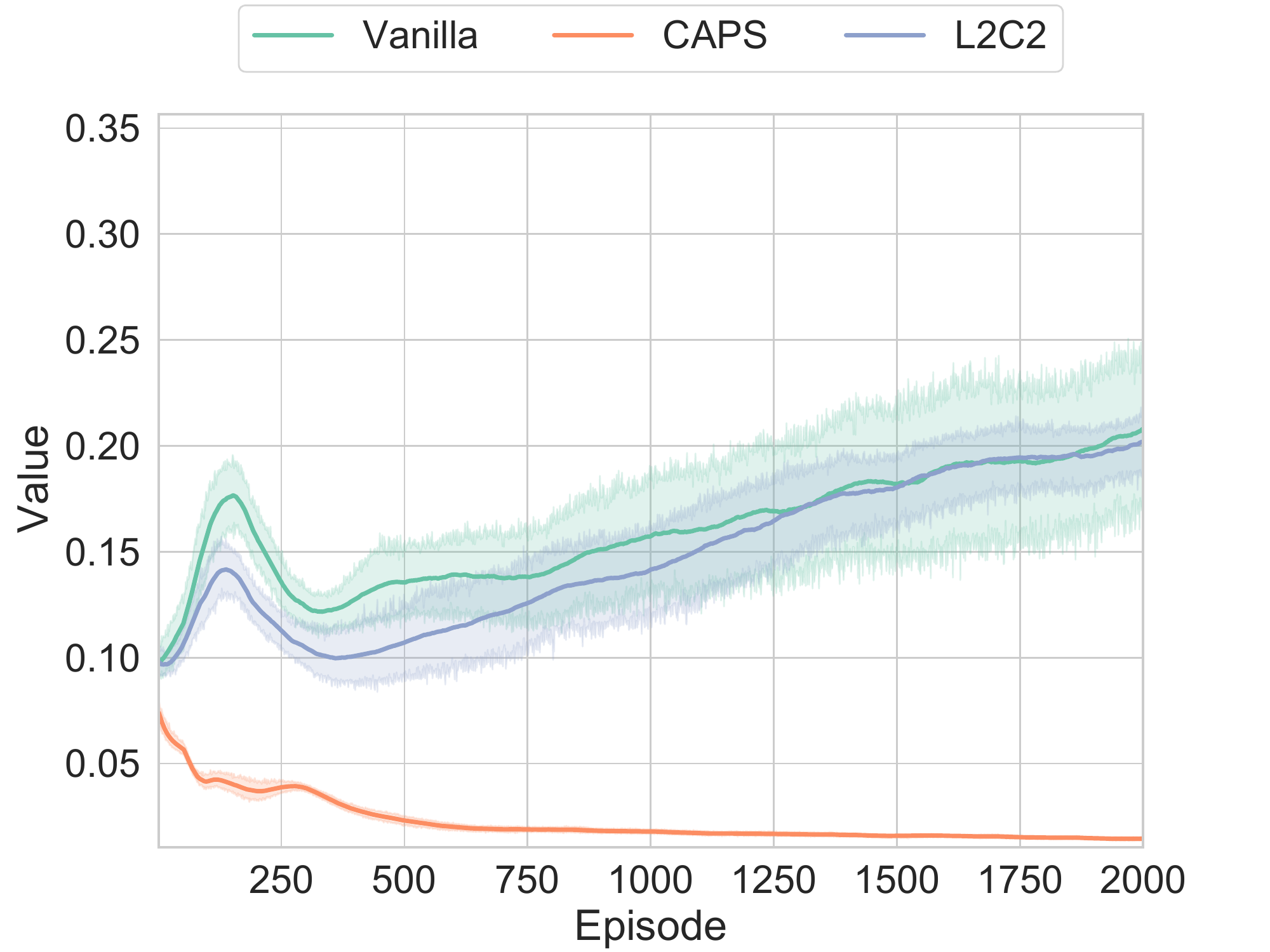}
        \subcaption{Hopper}
        \label{fig:smo_diff_action_Hopper}
    \end{subfigure}
    \begin{subfigure}[b]{0.24\linewidth}
        \centering
        \includegraphics[keepaspectratio=true,width=\linewidth]{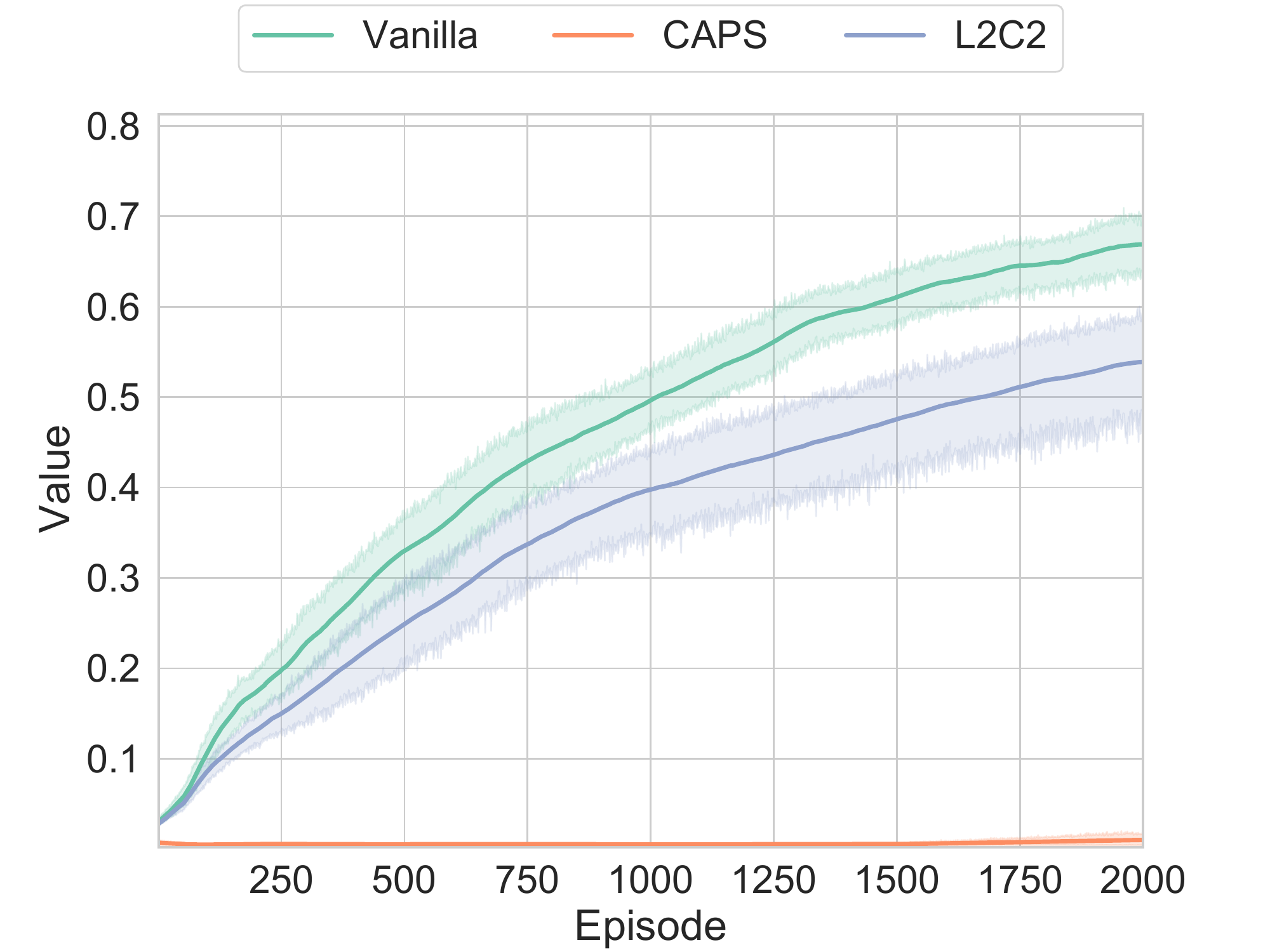}
        \subcaption{Ant}
        \label{fig:smo_diff_action_Ant}
    \end{subfigure}
    \caption{Learning curves for policy smoothness ($L_2$ norm of temporal actions):
    CAPS overly suppressed the change in policies;
    L2C2 basically improved the policy smoothness compared to Vanilla, while holding the expressiveness of function;
    for (a) Swingup, Vanilla made the policy smooth as the same level of smoothness as L2C2 after acquiring the swinging-up task;
    the policy smoothness of L2C2 in (c) Hopper seems to be deteriorated since its behavior would be global optimal different from the others with local optimal policy.
    }
    \label{fig:smo_diff_action}
\end{figure*}

\begin{table*}[tb]
    \caption{Test results:
        the numbers in parentheses denote standard deviations.
    }
    \label{tab:test}
    \centering
    \begin{tabular}{l cccc cccc}
        \hline\hline
         & \multicolumn{4}{c}{The sum of rewards}
         & \multicolumn{4}{c}{$L_2$ norm of temporal actions}
         \\
        Method & Swingup & Reacher & Hopper & Ant & Swingup & Reacher & Hopper & Ant
        \\
        \hline
        Vanilla
        & 649.57
        & 14.73
        & 1561.85
        & 2724.32
        & 0.084
        & 0.132
        & 0.157
        & 0.635
        \\
        (Without regularization)
        & (77.96)
        & (4.34)
        & (879.99)
        & (341.46)
        & (0.011)
        & (0.032)
        & (0.097)
        & (0.090)
        \\
        CAPS
        & 239.57
        & -1.57
        & 762.20
        & 911.36
        & 0.026
        & 0.003
        & 0.004
        & 0.002
        \\
        (Conventional method)
        & (26.56)
        & (1.53)
        & (321.26)
        & (98.33)
        & (0.002)
        & (0.001)
        & (0.001)
        & (0.001)
        \\
        L2C2
        & 693.77
        & 15.42
        & 2091.76
        & 2762.75
        & 0.075
        & 0.096
        & 0.174
        & 0.531
        \\
        (Proposed method)
        & (123.78)
        & (4.04)
        & (563.82)
        & (536.85)
        & (0.012)
        & (0.018)
        & (0.055)
        & (0.150)
        \\
        \hline
        L2C2
        & 707.41
        & 15.81
        & 1998.54
        & 2690.26
        & 0.075
        & 0.095
        & 0.141
        & 0.521
        \\
        (Ablation test: $\beta=0$)
        & (40.94)
        & (4.196)
        & (714.88)
        & (483.62)
        & (0.010)
        & (0.023)
        & (0.063)
        & (0.142)
        \\
        \hline\hline
    \end{tabular}
\end{table*}

The learning curves for the score and the policy smoothness with the above setups are shown in Figs.~\ref{fig:score} and~\ref{fig:smo_diff_action}.
The test results with the respective learned policies are summarized in Table~\ref{tab:test}.
Note that L2C2 with $\beta=0$ was also conducted as an ablation test and is listed in Table~\ref{tab:test}.

Obviously, the conventional method, CAPS, fell into local solutions for every task.
This is because the policy smoothness converges to a very small value only for CAPS in Fig.~\ref{fig:smo_diff_action}, which means that CAPS made the policy excessively smooth.
This lost the capability to handle scenes that require changes in the policy (i.e. expressiveness), such as a switch from a swing-up motion to an inverted balance control in Swingup.
Note that this problem may be able to be avoided by choosing the appropriate regularization gains and noise scale, but they are problem-dependent, losing applicability of CAPS.

The remaining results with and without L2C2 are compared.
As summary, in terms of task performance, the learning curves of L2C2 are generally consistent with that of Vanilla.
In addition, the proposed L2C2 outperformed Vanilla in the test results.
In details, for Hopper, the task performance of L2C2 was greatly improved, probably because the policy smoothness suppressed the risky exploration that would cause the failures.
On the other hand, the suppression of exploration caused delay or failure in learning for Ant with the highest-dimensional action space.
Thus, while L2C2 retains the expressiveness necessary for the task acquisition, it suppresses the exploratory performance of the learning process.
However, as mentioned above, excessive exploration is risky and consumes a lot of energy, hence this suppression would be useful in terms of real robotic applications.

By comparing the policy smoothness, it is found that L2C2 basically improved it from the early stage of learning.
Note that the policy smoothness for Hopper is meaningless because the acquired behavior learned by L2C2 (i.e. the globally optimal policy) would be different from one learned by Vanilla (i.e. the locally optimal policy).
In addition, compared to CAPS, which impairs the task performance, L2C2 does not have sufficiently smooth policies.
It may be required to maximize the locally compact space without degrading the task performance.

Finally, the ablation test of $\beta=0$ to eliminate the regularization of the value function is compared.
The test results show that there is not much difference from L2C2 with $\beta=0.1$, suggesting that the regularization of the value function in L2C2 is not effective in this implementation.
One of the reasons for this would be that the training signals generated by the value function are sufficiently stabilized by the introduction of the target network and ensemble learning.
Indeed, when $\tau$ in t-soft update is increased from $0.1$ to $0.5$, the score for L2C2 with $\beta=0.1$ was 376, while it dropped to 284 for L2C2 with $\beta=0$.
In addition, only by L2C2 with $\beta=0.1$, the score more than 800, which indicates the success of Swingup, was obtained.

From the above results, it can be concluded that the proposed L2C2 smoothes the policy and value functions without deteriorating the expressiveness required for the task, and contributes to the improvement of learning performance.
It is also confirmed that the suppression of exploration performance occurs as a side effect, but it should be noted that the effectiveness of this is task-dependent.

\section{Summary and future work}

In this paper, a new regularization method for RL, so-called L2C2, was proposed.
According to the definition of local Lipschitz continuity, L2C2 was first derived as a hard-constrained minimization problem.
Such a hard constraint cannot be handled directly due to NP-hard.
Therefore, L2C2 was converted into the regularization terms via Lagrange multiplier, so that it makes the policy and value functions smooth locally and not lose the expressiveness.
The neighborhood of the current state was designed based on the travel distance of state transition, which implicitly evolved the spatially-local regularization into the spatio-temporal one.

To verify the proposed L2C2, four numerical noisy simulations were performed as benchmarks.
As a result, L2C2 achieved the equivalent or better task performance compared to the conventional methods without losing the expressiveness of the policy.
In addition, L2C2 succeeded in smoothing the policy better than the case without regularization.

Although L2C2 has indeed improved the RL performance for more practical robotic situations (i.e. noisy observation) compared to the previous methods, there is still room for improvements in the implementation of its regularization way.
For example, in eqs.~\eqref{eq:reg_policy} and~\eqref{eq:reg_value}, the original definition of the upper bound is replaced as the expected value due to the difficulty in computing it.
To utilize the upper bound as it is, $\tilde{s}$ should be generated adversarially, referring to SA-MDP~\cite{zhang2020robust,zhang2021robust}, although it will increase the training cost of the adversarial networks.
In addition, the hard constraints were replaced by a linear weighted sum, which fails to obtain a non-convex Pareto solution when viewed as a multi-objective optimization problem.
In order to satisfy the requirements of the minimization problem as much as possible, an alternative formulation, which can obtain any Pareto solutions (e.g. Tchebycheff scalarization~\cite{steuer1983interactive,aotani2021meta}), should be required to replace the linear weighted sum.
These improvements in the near future would make L2C2 applicable to real robot applications.

%
%
%
\section*{ACKNOWLEDGMENT}

This work was supported by JSPS KAKENHI, Grant-in-Aid for Scientific Research (B), Grant Number JP20H04265.

\bibliographystyle{IEEEtran}
{
\bibliography{biblio}
}

\end{document}